\documentclass{llncs}
\usepackage{cite}
\usepackage{amsmath,amssymb,amsfonts}
\usepackage{breqn}

\usepackage{graphicx}
\usepackage{textcomp}
\usepackage{xcolor}
\usepackage{multirow}
\usepackage{pgf-umlsd}
\usepackage{comment} 
\usepackage{url}
\usepackage{geometry}
\usepackage[linesnumbered,ruled,vlined]{algorithm2e}
\SetKwInput{KwInput}{Input}
\SetKwInput{KwOutput}{Output}
\SetKwComment{Comment}{/* }{ */}
\usepackage[linesnumbered,ruled,vlined]{algorithm2e}
\SetKwInput{KwInput}{Input}

\begin{document}

\title{FL-Defender: Combating Targeted Attacks in Federated Learning}

%\thanks{Supported by organization x.}}
%
\titlerunning{FL-Defender: Combating Targeted Attacks in Federated Learning}
%JOSEP2. Supressed authors for blind submission
\author{Najeeb Jebreel \and
Josep Domingo-Ferrer
}

\authorrunning{N. Jebreel et al.}

\institute{Universitat Rovira i Virgili,\\ Department of Computer Engineering and Mathematics,\\
CYBERCAT-Center for Cybersecurity Research of Catalonia,\\ UNESCO Chair in Data
Privacy,\\ Av. Pa\"isos Catalans 26, 43007 Tarragona, Catalonia\\
\email{\{najeeb.jebreel, josep.domingo\}@urv.cat}}

\maketitle              % typeset the header of the 
\begin{abstract} 
Federated learning (FL) enables learning a global machine learning model from local data distributed among a set of participating workers. 
This makes it possible i) to train more accurate models due to learning from rich joint training data, and ii) to improve privacy by not sharing the workers' local private data with others. 
However, the distributed nature of FL makes it vulnerable to targeted poisoning attacks that negatively impact the integrity of the learned model while, unfortunately, being difficult to detect. 
Existing defenses against those attacks are limited by assumptions on the workers' data distribution, may degrade the global model performance on the main task and/or are ill-suited to high-dimensional models.
In this paper, we analyze targeted attacks against FL and find that the neurons in the last layer of a deep learning (DL) model that are related to the attacks exhibit a different behavior from the unrelated neurons, making the last-layer gradients valuable features for attack detection. 
Accordingly, we propose \textit{FL-Defender} as a method to combat FL targeted attacks. It consists of 
i) engineering more robust discriminative features by calculating the worker-wise angle similarity for the workers' last-layer gradients, ii) compressing the resulting similarity vectors using PCA to reduce redundant information, and iii) re-weighting the workers' updates based on their deviation from the centroid of the compressed similarity vectors. 
Experiments on three data sets with different DL model sizes and data distributions show the effectiveness of our method at defending against label-flipping and backdoor attacks. Compared to several state-of-the-art defenses, FL-Defender achieves the lowest attack success rates, maintains the performance of the global model on the main task and causes minimal computational overhead on the server.

\keywords{Federated learning \and Security and robustness \and Targeted poisoning attacks  \and Label-flipping attacks  \and Backdoor attacks}
\end{abstract}

\section{Introduction}

%JOSEP2. Slightly rewritten.
Federated learning (FL)~\cite{mcmahan2017communication, konevcny2015federated} enables building an ML model from private data distributed among a set of participating worker devices. In FL, the workers fine-tune a global model received from the server on their local data to compute local model updates that they upload to the server, which aggregates them to obtain an updated global model. This process is iterated until it converges to a high-quality global model.
Therefore, FL improves accuracy, privacy and scalability, respectively, by learning from big joint data, keeping the workers' local data at their respective devices ({\em e.g.} smartphones) and distributing the training load across the workers' devices~\cite{bonawitz2019towards}.

%JOSEP2. Slightly rewritten.
Despite these benefits, the distributed nature of FL makes it vulnerable to poisoning attacks~\cite{blanco2021achieving,ferrag2021federated}. 
Since the server cannot control the behavior of the workers, any of them may deviate from the prescribed training protocol to attack the model by conducting either untargeted poisoning ({\em i.e.,} Byzantine) attacks~\cite{blanchard2017machine, wu2020federated} or targeted poisoning attacks~\cite{biggio2012poisoning, fung2020limitations, bagdasaryan2020backdoor}. 
In the former attacks, the attacker aims at degrading the model's overall performance; in the latter attacks, he aims at causing the global model to incorrectly classify some attacker-chosen inputs. 
Furthermore, poisoning attacks can be performed in two ways: model poisoning~\cite{blanchard2017machine, wu2020federated, bagdasaryan2020backdoor} or data poisoning~\cite{biggio2012poisoning,fung2020limitations,tolpegin2020data}. 
In model poisoning, the attackers maliciously manipulate their local model parameters before sending them to the server. 
In data poisoning, they inject fabricated or falsified data examples into their training data before local model training. 
Both attacks result in poisoned updates being uploaded to the server, in order to prevent the global model from converging or to bias it. Compared to untargeted poisoning~\cite{blanchard2017machine, wu2020federated}, targeted attacks~\cite{biggio2012poisoning,fung2020limitations,tolpegin2020data, bagdasaryan2020backdoor,wang2020attack, xie2019dba} are more serious, given their stealthy nature and severe security implications~\cite{steinhardt2017certified, fung2020limitations,tolpegin2020data, awan2021contra}.

Several defenses against poisoning attacks have been proposed, which we discuss in Section \ref{related}.
However, they are either impractical~\cite{nelson2008exploiting, jagielski2018manipulating}, may degrade the performance of the aggregated model on the main task~\cite{bagdasaryan2020backdoor, wang2020attack} or assume specific distributions of local training data~\cite{shen2016auror,blanchard2017machine,tolpegin2020data,fung2020limitations, awan2021contra,chen2017distributed,yin2018byzantine}. Specifically,
%JOSEP. I assume "poses" means "possess". I rewrite a bit what follows.
\cite{jagielski2018manipulating} assumes the server to possess some data examples representing the workers' data, which is not a realistic assumption in FL;
\cite{shen2016auror,blanchard2017machine,tolpegin2020data,chen2017distributed,yin2018byzantine} assume the local data are independent and identically distributed (iid) among workers, which leads to poor performance  on non-iid data~\cite{awan2021contra}; 
\cite{fung2020limitations,awan2021contra} identify workers with  similar updates as attackers, which leads to wrongly penalizing honest workers with similar local data~\cite{nguyen2021flguard, li2021auto}.
Besides those assumptions, the dimensionality of the model is a paramount factor that affects the performance of most of the methods above: high-dimensional models are more vulnerable to targeted poisoning attacks because they cause small changes on a poisoned local update without being detected~\cite{chang2019cronus}. 
To the best of our knowledge, no contribution to the state of the art provides an effective defense against targeted poisoning attacks without being hampered by the data distribution or model dimensionality.

%JOSEP. A bit rewritten.
{\bf Contributions and plan.}
In this paper, we study targeted poisoning attacks by analyzing the behavior of label-flipping (LF) and backdoor (BA) attacks on DL models. We find that an unbalanced distribution of the workers' local data and a high dimensionality of the DL model make the detection of these attacks quite challenging.
Moreover, we observe that attack-related last-layer neurons exhibit a different behavior from attack-unrelated last-layer neurons, which makes the last-layer gradients useful features for detecting targeted attacks.
Accordingly, we propose \textit{FL-Defender}, a method that can mitigate the attacks regardless of the model dimensionality or the distribution of the workers' data. 
First, we use the workers' last-layer gradients to engineer more robust discriminative features that capture the attack behavior and discard redundant information.
Specifically, we compute the worker-wise angle similarity for the workers' last-layer gradients and then compress the computed similarity vectors using principal component analysis (PCA)~\cite{wold1987principal} to reduce redundant information. After that, we penalize the workers' updates based on their angular deviation from the centroid of the compressed similarity vectors.
Experimental results on three data sets with different DL model sizes and worker data distributions demonstrate the effectiveness of our approach at defending against the attacks. Compared with several state-of-the-art defenses, \textit{FL-Defender} achieves better performance at retaining the accuracy of the global model on the main task, reducing the attack success rate and causing minimal computational overhead on the server. 

The rest of this paper is organized as follows. 
Section~\ref{prelim} introduces preliminary notions. 
Section~\ref{lf_attack_threat_model} formalizes the attacks and the threat model being considered. 
Section~\ref{related} discusses countermeasures for poisoning attacks in FL.
Section~\ref{attacks_analysis} analyzes the behavior of label-flipping and backdoor attacks, and shows the robustness of the engineered features.
Section~\ref{meth} presents the methodology of the proposed defense. 
Section~\ref{evaluation} details the experimental setup, and reports the obtained results. 
Finally, conclusions and future research lines are gathered in Section~\ref{conclusion}.

\section{Preliminaries}
\label{prelim}

\textbf{Deep neural network-based classifiers.}
A deep neural network (DNN) is a function $F(x)$, obtained by composing $L$ functions $f^l, l\in [1, L]$, that transforms an input $x$ to a predicted output $\hat{y}$. Each $f^l$ is a layer that is parameterized by a weight matrix $w^l$, a bias vector $b^l$ and an activation function $\sigma^l$. $f^l$ takes as input the output of the previous layer $f^{l-1}$. The output of $f^l$ on an input $x$ is computed as $f^l(x) = \sigma^l(w^l \cdot x + b^l)$. Therefore, a DNN can be formulated as \[    F(x) = \sigma^L(w^L \cdot \sigma^{L-1}(w^{L-1} \dots \sigma^1(w^1 \cdot x + b^1) \dots + b^{L-1} )+ b^L).\]
A DNN-based classifier consists of a feature extractor and a classifier~\cite{krizhevsky2017imagenet, minaee2021deep}. 
The classifier makes the final classification decision based on the extracted features and usually consists of one or more fully connected layers where the last layer contains $|\mathcal{C}|$ neurons with $\mathcal{C}$ being the set of all possible class values. 
The output layer's vector $o \in \mathbb{R^{|\mathcal{C}|}}$ is usually passed to the softmax function that converts it to a vector $p$ of probabilities, which is called the vector of confidence scores.

In this paper, we use predictive DNNs as $|\mathcal{C}|$-class classifiers, where the index of the highest confidence score in $p$ is considered the final predicted class $\hat{y}$. 
Also, we analyze the last layers' gradients of DNNs to filter out poisoned updates resulting from LF and backdoor attacks.

\textbf{Federated learning.}
In FL, an aggregator server and $K$ workers cooperatively build a shared global model. 
The server starts by randomly initializing the global model $W^t$. 
Then, at each training round, the server selects 
a subset of workers $S$ of size $C \cdot K \ge 1$
where $K$ is the total number of workers in the system, and $C$ is the fraction of workers that are selected in the training round. After that, the server distributes the current global model 
%JOSEP2. Changed w^t to W^t. Slight rewritten below.
$W^t$ to all workers in $S$. 
Besides the global model, the server sends a set of hyper-parameters to be used at the workers' side to train their model locally: number of local epochs $E$, local batch size $BS$ and learning rate $\eta$. 
After receiving the new shared model $W^t$, each worker divides her local data into batches of size $BS$ and performs $E$ local training epochs of stochastic gradient descent (SGD). Finally, workers upload their updated local models $W_{(k)}^{t+1}$ to the server, which then aggregates them to obtain the new global model $W^{t+1}$. 
The federated averaging algorithm (\textit{FedAvg})\cite{mcmahan2017communication} is usually employed to perform the aggregation as 
\[   W^{t+1} = \sum_{k=1}^{K} \frac{n_{(k)}}{n} W^{t+1}_{(k)},\]
%JOSEP2. Rewritten.
where $n_{(k)}$ is the number of data points locally held by worker $k$ and $n$ is the total number of data points locally held by the $K$ workers, that is, $n=\sum_{k=1}^{K} n_{(k)}$.  

\section{Attacks and threat model}
\label{lf_attack_threat_model}

%JOSEP. Slightly rewritten.
As mentioned above, we focus on the LF and BA attacks, two widely used targeted attacks in the FL literature.
In the LF attack~\cite{biggio2012poisoning,fung2020limitations}, each attacker poisons his training data set $D_k$ as follows: for all examples in $D_k$ with a class label $c_{src}$, change their class label to $c_{target}$. An example is changing the labels of ''fraudulent'' activities to ''non-fraudulent''.
In the BA attack~\cite{gu2017badnets}, the attacker poisons his training data by embedding a specific pattern (the backdoor) into training examples with specific features and assigns them a target class label of his choice. 
The pattern acts as a trigger for the global model to output the desired target label for the backdoored examples. 
As a famous example, the attacker could put a small sticker on a ''stop traffic sign'' and change its label to ''speed limit''.
After poisoning their data, the attacker trains his local models on the poisoned data, with the same training settings as the honest workers, and uploads the resulting poisoned updates to the server to be aggregated into the global model.

\textbf{Assumptions on training data distribution.} 
Since the local data sets of the workers may come from heterogeneous sources~\cite{bonawitz2019towards,wang2019edge}, they may be either identically distributed (iid) or non-iid. In the iid setting, each worker holds local data representing the whole distribution. 
In the non-iid setting, the distributions of the workers' local data sets can be different in terms of the classes represented in the data and/or the number of samples each worker holds for each class.
Consequently, each worker may have local data with i) all the classes being present in a similar proportion as in the other workers' local data (iid setting), ii) some classes being present in a different proportion (non-iid setting). 

%JOSEP2. Slightly rewritten.
\textbf{Threat model.}
We consider a number of attackers $K'\leq K/5$, that is, no more than $20\%$ of the $K$ workers in the system. 
Although some works in the literature assume larger percentages of attackers, finding more than 20\% of attackers in real-world FL scenarios is unlikely.
For example,  with millions of users~\cite{davenport_corbin} in Gboard~\cite{hard2018federated}, controlling a small percentage of user devices requires the attacker(s) to compromise a large number of devices, which demands huge  effort and resources and is therefore impractical. 
Furthermore, we assume the FL server to be honest and not compromised, and the attackers to have no control over the aggregator or the honest workers.
The attacker's goal for the LF attack is to cause the learned global model to classify the source class  examples into the target class at test time.
The attacker's goal for the BA attack is to fool the global model into falsely predicting the attacker's chosen class for any target example carrying the backdoor pattern, while maintaining the benign model performance on non-backdoored examples.

\section{Related work}
\label{related}

Existing methods to counter poisoning attacks in FL are based on one of the following principles.

\textbf{Evaluation metrics.} An update is penalized as being probably bad if it degrades an evaluation metric of the global model, {\em e.g.} its accuracy. \cite{nelson2008exploiting, jagielski2018manipulating} use a validation set on the server to compute the loss caused by each local update and then keep or discard an update based on its computed loss. However, this is impractical in FL because the server does not have access to the workers' data.
In addition, this approach cannot properly detect backdoor attacks because these attacks have little to no impact on the model's performance on the main task.

\textbf{Clustering updates.} Updates are clustered into two separate groups, where the smaller group contains all the bad updates, that are subsequently disregarded in the model learning process. The 
%JOSEP2. Slightly rewritten.
Auror~\cite{shen2016auror} and the multi-Krum (MKrum)~\cite{blanchard2017machine} methods assume that the workers' data are iid, which explains their poor performance on non-iid data~\cite{awan2021contra}, where they incur high false positive and false negative rates.

\textbf{Worker behavior.} This approach assumes the attackers' behavior is reflected in their updates, making them more similar to each other than honest workers' updates. Hence, updates are penalized based on their similarity. For example, FoolsGold~\cite{fung2020limitations} and CONTRA~\cite{awan2021contra} limit the contributions of similar updates by reducing their learning rates or preventing their originators from being selected. However, these methods also penalize similar good updates, which results in significant drops in the model performance~\cite{nguyen2021flguard, li2021auto} when good updates are similar. 

\textbf{Update aggregation.} This approach uses robust update aggregation rules, such as the median~\cite{yin2018byzantine}, the trimmed mean~\cite{yin2018byzantine} or the repeated median~\cite{siegel1982robust}. However, the performance of these rules 
%JOSEP. Rewritten.
deteriorates on non-iid data because they discard most of the information at the time 
of update aggregation. Also, their estimation error scales up proportionally to the square root of the model size~\cite{chang2019cronus}.

\textbf{Differential privacy (DP).} Methods under this approach~\cite{bagdasaryan2020backdoor, sun2019can} clip individual update parameters to a maximum threshold and add random noise to the parameters to reduce the impact of potentially poisoned updates on the aggregated global model. 
However, there is a trade-off between adding noise to mitigate the attacks and maintaining the benign performance of the aggregated model on the main task~\cite{bagdasaryan2020backdoor, wang2020attack}. 
Also, DP-based methods consider only mitigating backdoor attacks and are not designed to counter label-flipping attacks.

Several works propose to analyze specific parts of the updates to counter poisoning attacks. 
\cite{jebreel2020efficient} proposes analyzing the last layer's biases. However, it assumes the local data are iid and form two separate clusters. 
FoolsGold~\cite{fung2020limitations} analyzes the last layer's weights to counter targeted poisoning attacks. However, as mentioned above, it performs poorly when good updates are similar because it considers them to be bad. To counter LF attacks, \cite{tolpegin2020data} uses PCA to analyze the weights associated with \textit{the possibly attacked source class} and excludes potential bad updates that differ from the majority of updates in those weights. 
%JOSEP2. A bit rewritten.
However, the method is evaluated only for LF attacks under the iid setting, and it requires prior knowledge on the possible source class. 

The methods just cited share the shortcomings of (i) making assumptions on the distributions of the workers' data and (ii) not providing analytical or empirical evidence of why focusing on specific parts of the updates contributes towards defending against the attacks.
In contrast, we provide comprehensive conceptual and empirical analyses that explain why focusing on the last-layer gradients is especially useful to defend against targeted poisoning attacks. 
Also, we propose a more robust method that can mitigate the attacks without being limited by the distribution of the workers' data or the dimension of the models in use.

\section{Analysing targeted attacks against FL}
\label{attacks_analysis}

%JOSEP2. A bit rewritten.
This section is key in our work. We study the behavior of label-flipping and backdoor attacks to find robust discriminative features that can detect such attacks.

Let us consider an FL classification task where each local model is trained with the cross-entropy loss over one-hot encoded labels as follows.
First, the activation vector $o$ of the last layer neurons (a.k.a. logits) is fed into the softmax function to compute the vector $p$ of probabilities as follows:
%\begin{equation}
%    \label{softmax}
%JOSEP. Added range of subscript and changed from i to k
\[         p_k = \frac{e^{o_k}}{\sum_{j = 1}^{|\mathcal{C}|} e^{o_j}},\;\;\; k=1, \ldots, |\mathcal{C}|.\]
%\end{equation}
Then, the loss is computed as
%\begin{equation}
%    \label{loss_cross_entropy}
\[    \mathcal{L}(y, p) = - \sum_{k= 1}^{|\mathcal{C}|} y_k \log(p_k),\]
%\end{equation}
where $y = (y_1, y_2, \ldots, y_{|\mathcal{C}|})$ is the corresponding one-hot encoded true label and $p_k$ denotes the confidence score predicted for the $k^{th}$ class. 
After that, the gradient of the loss w.r.t. the output $o_{i}$ of the $i^{th}$ neuron (a.k.a the $i^{th}$ neuron error)  in the output layer is computed as 
\begin{equation}
\label{derivative_neuron}
    \delta_i = \frac{\partial \mathcal{L}(y, p)}{\partial o_i}
    = p_i - y_i.
\end{equation}
Note that $\delta_i$ will always be in the interval $[0, 1]$ when $y_i = 0$ (for the wrong class neuron), while it will always be in the interval $[-1, 0]$ when $y_i = 1$ (for the true class neuron).

The gradient $\nabla b^L_i$ w.r.t. the bias $b^L_i$ connected to the $i^{th}$ neuron in the output layer can be written as 
\begin{dmath}
\label{derivative_bias}
    \nabla b^L_i  =  \delta_i \frac{\partial \sigma^L}{\partial (w^L_i \cdot a^{L-1} + b^L_{i})},
\end{dmath}
where $a^{L-1}$ is the activation output of the previous layer.
Likewise, the gradient $\nabla w^L_i$ w.r.t. the weights vector $w^L_i$ connected to the $i^{th}$ neuron in the output layer is
\begin{dmath}
\label{derivative_weights}
    \nabla w^L_i =  \delta_i  a^{L-1} 
    \frac{\partial \sigma^L}{\partial (w^L_i \cdot a^{L-1} + b^L_{i})}. 
\end{dmath}

From Equations~\eqref{derivative_bias} and~\eqref{derivative_weights}, we can notice that $\delta_i$ directly and highly impacts on the gradients of the output layer's weights and biases. \\

%A bit rewritten.
\textbf{Behavior of label-flipping attacks.} 
In FL, a label-flipping attacker always tries to minimize $p_{c_{src}}$ for any example in his training data, including examples that belong to class $c_{src}$. 
On the other side, he always tries to maximize $p_{c_{target}}$ for examples that belong to $c_{src}$ during model training.
Since this goes in the opposite direction of the objective of honest workers for examples in $c_{src}$, the attack will entail substantial alteration of
$\delta_{c_{src}}$ and $\delta_{c_{target}}$, as it can be seen from Expression \eqref{derivative_neuron}.
In turn, from Expressions \eqref{derivative_bias} and 
\eqref{derivative_weights}, it follows that 
altering $\delta_{c_{src}}$ and $\delta_{c_{target}}$ directly alters the biases and weights corresponding to the output neurons of $c_{src}$ and $c_{target}$ during the training of the attacker's local model.
Hence, the impact of the attack can be expected to show in the gradients of the last-layer neurons corresponding to 
$c_{src}$ and $c_{target}$.
%JOSEP2. IMPORTANT. Rewritten. Please check.
%Najeeb. It's fine. I just made a slight addition.
However, the last layer is likely to contain other neurons unrelated to the attack where both the attacker and the honest workers share the same objectives, which makes the attack harder to spot. 
Considering all layers is still worse, because in the layers different from the last one the impact of the attack will be even less perceptible 
%JOSEP3. Rewritten.
(as it will be mixed with more unrelated parameters).
%Najeeb. Added "because it will be mixed by more unrelated parameters".
%because it will be mixed by more unrelated parameters.
%This applies even more to the whole model gradients (with all layers), because the impact of the attack will  among many unaffected unrelated gradients. Thus, analyzing all the update layers yields worse detection than focusing on the last layer.   
Moreover, there are two other factors that increase the difficulty of detecting the attack by analyzing the update as a whole: i) the early layers usually extract common features that are not class-specific~\cite{nasr2019comprehensive} and ii) in general, most parameters in DL models are redundant \cite{denil2013predicting}. That causes the magnitudes and angles of the bad and good updates to be similar, which makes models with large dimensionality an ideal environment for a successful label-flipping attack.

%JOSEP. A bit rewritten.
\textbf{Behavior of backdoor attacks.} 
Backdoor attacks might be viewed as a particular case of label-flipping attacks because the attacker flips the label of a training example when it contains a specific feature or pattern, whereas he retains the correct label when the example does not contain the pattern (clean example). 
However, since the global model will correctly learn from a majority of honest workers and, in the clean examples, from the attackers as well, the received global model will probably overlook the backdoor pattern and assign the correct classes to backdoored examples, especially in the early training iterations.
This will prompt the attackers to try to minimize $p_{c_{src}}$ and maximize $p_{c_{target}}$ for the backdoored examples, which can be expected to stand out in the magnitudes and the directions of the gradients contributed by the attackers.
On the other hand, since the attackers also try to maximize $p_{c_{src}}$ for their clean examples, the impact of the backdoor attacks is expected to be stealthier compared to that of LF attacks even when looking at the last-layer gradients.

%JOSEP2. A bit rewritten.
\textbf{Engineering more robust features to detect attacks.} 
From the above analysis, it is clear that focusing on analyzing last-layer gradients is more helpful to detect targeted attacks than analyzing all layers. Nevertheless, the presence of a large number of attack-unrelated gradients in the last layer may still render attack detection difficult.
Getting rid of those redundant and unrelated gradient features could help obtain more robust discriminatory features for targeted attacks.
Since the impact of the targeted attacks is directly reflected in the directions of gradients of attack-related neurons, comparing the difference in directions between the gradients of a good update and a poisoned update can be expected to better capture the attack's behavior. 
If we look at the angular similarity of the workers' last-layer gradients, the similarity values between good and poisoned updates are expected to display unique characteristics in the computed similarity matrix. 
PCA can be used to capture those unique characteristics from the matrix and reduce redundant features.

\textbf{Empirical analysis.}
To empirically validate our previous conceptual discussion, we used $20$ local updates resulting from simulating an FL scenario under the LF and BA attacks with each of the CIFAR10-IID  and CIFAR10-non-IID benchmarks, where $4$ updates (that is, 20\%) were poisoned. In these two benchmarks, the ResNet18~\cite{he2016deep} architecture, which contains about $11M$ parameters was used. In addition, training data were randomly and uniformly distributed among workers in CIFAR-IID, while we adopted a Dirichlet distribution~\cite{minka2000estimating} with $\alpha = 1$ to generate non-iid data for the $20$ workers in the CIFAR10-non-IID.
The details of the experimental setup are given in Section~\ref{setup}.

%JOSEP2. IMPORTANT. Rewritten. It was hard for me to understand
%what exactly was done. I have tried to write it more clearly for me,
%but I'm not sure whether this is what you really did.
%Najeeb. It's fine. This is exactly what I did.
Then, for each benchmark and attack scenario, we computed the following:
\begin{itemize}
    \item The first two principal components (PCs) of the all-layer gradients for each local update. Then, we computed the centroid (CL) of the first two PCs for the $20$ local updates. After that, we computed the angle between CL and every pair of PCs for each update.
    
    \item  The centroid (CL) of the last-layer gradients for the $20$ local updates and the angle between CL and each last-layer gradient for each update. 
    
    \item The first two PCs of the last-layer gradients for each local update. Then, we computed the centroid (CL) of the first two PCs of the $20$ last-layer gradients. After that, we computed the angle between CL and every pair of PCs for each last-layer gradient.
    
    \item The cosine similarity for the $20$ workers' last-layer gradients. Then, we computed the first two PCs of the similarity matrix. After that, we computed the centroid (CL) of the first two PCs of the $20$ similarity vectors. 
    Finally, we computed the angle between CL and every pair of PCs of each similarity vector.
    
\end{itemize}

Once the above computations were completed, we visualized the magnitude of each input, and the angle between the input and its corresponding centroid.

\begin{figure}[!htbp]
      \centering
      \includegraphics[width=0.8\linewidth]{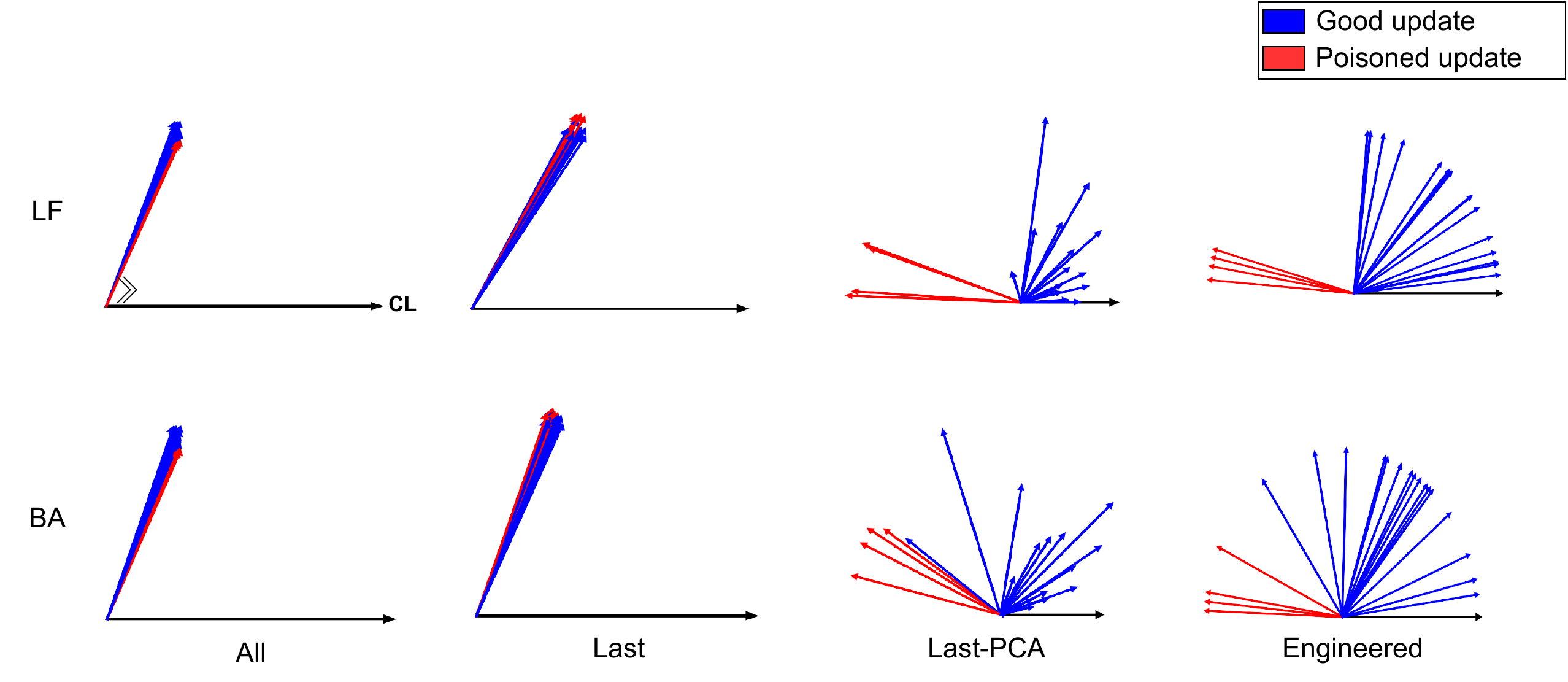}
\caption{Deviation of CIFAR10-IID gradient features from the centroid}
\label{fig:cifar10_iid}
\end{figure}

Fig.~\ref{fig:cifar10_iid} shows the visualized vectors for the CIFAR-IID benchmark. For the LF attack, we can see that analyzing the first two PCs of the all-layer gradients (All) led to poisoned updates and good updates with very similar magnitudes and angular deviation from the centroid, which made it quite challenging to tell them apart. The same applies to analyzing the last-layer gradients (Last).
On the other hand, analyzing the first two PCs of the last-layer gradients (Last-PCA) led to an apparent separation between good and bad updates. This also applied to analyzing the engineered features (Engineered).
For the BA attack, the results were similar with the difference that our engineered features allowed better separation than Last-PCA.
This confirms our conceptual discussion that redundant and attack-unrelated gradients make the attacks stealthier.
\vspace{-3ex}

\begin{figure}[!htbp]
      \centering
      \includegraphics[width=0.8\linewidth]{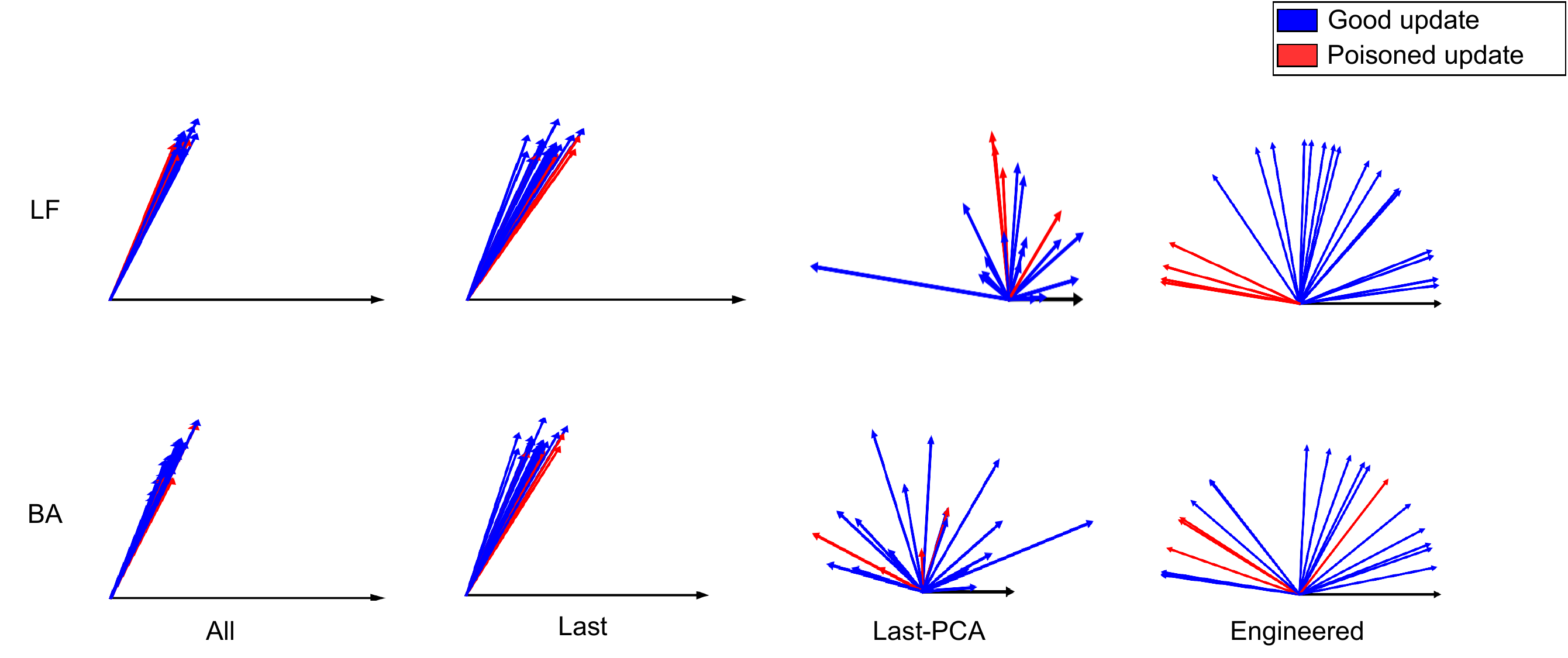}
\caption{Deviation of CIFAR10-non-IID gradient features from the centroid}
\label{fig:cifar10_noniid}
\end{figure}

\vspace{-3ex}
Fig.~\ref{fig:cifar10_noniid} shows the visualized vectors for the CIFAR-non-IID benchmark, where the data were non-iid among workers. 
For the LF attack, we can see that only our engineered features provided robust discrimination between good and bad updates.
For the BA attack, even if our engineered features allowed better separation, it was challenging to tell updates apart.
This is because of the impact of the non-iidness and also because BA attacks are stealthier than LF attacks.
This again confirms our intuitions and shows that our engineered features are more useful than the alternatives to detect targeted attacks.
\vspace{-3ex}

\vspace{-3ex}
%JOSEP2. Rewritten.
We were able to conclude from these analytical and empirical explorations that a high model dimensionality and the distribution of the workers' training data highly impact on the ability to discriminate between good updates and updates poisoned by targeted attacks.
Fortunately, it also became evident that we can use the last-layer gradients to engineer more robust discriminative features for attack
detection.

\section{FL-Defender design}
\label{meth}

%JOSEP2. Rewritten.
In this section, we present the design of FL-Defender, our proposed defense against FL targeted poisoning attacks. Our aims are:
1) to prevent attackers from achieving their goals by mitigating the impact of their poisoned updates on global model, 
2) to maintain the global model performance on the main task, 
3) to stay robust against the attacks regardless of the model size or the workers' data distribution, and 
4) to avoid substantially increasing the computational cost at the server. 
According to the lessons learned in the previous section, our defense first extracts the last-layer gradients of the workers' local updates, computes the worker-wise cosine similarity and then compresses the computed similarity vectors using PCA to reduce redundant information and extract more robust features.
After that, it computes the centroid of the compressed similarity vectors and computes the cosine similarity values between the centroid and each compressed vector.
The similarity values are accumulated during training and used, after re-scaling them, to re-weight the workers updates in the global model aggregation.

We formalize our method in Algorithm~\ref{algorithm1}. 
The aggregator server $A$ starts a federated learning task by  initializing the global model $W^{0}$ and the history vector $H^0$ that is used to accumulate the similarities between the directions of the workers' engineered features and their centroid.

Then, in every training iteration $t$, $A$ selects a random subset $S$ of $m$ workers and sends the current global model $W^t$ to the $m$ selected workers. 
Each worker $k\in S$ locally trains $W^{t}$ on her data $D_k$ and sends her local update $W_{k}^{t+1}$ back to $A$. 
Once $A$ receives the $m$ local updates, it separates the gradients of the last layers to obtain the set $\{\nabla L^{t+1}_k|k \in S\}$.

\vspace{0cm}
\begin{algorithm}[!ht]
\SetKwProg{Fn}{Function}{}{end}
\caption{FL-Defender: Combating targeted attacks in FL}
\label{algorithm1}
\SetAlgoLined
\KwInput{$K, C, BS, E, \eta, T$}
\KwOutput{$W^T$, the global model after $T$ training rounds}

$A$ initializes $W^{0}$,  $H^0 = \{H_k^0 = 0\}_{k = 1}^{K}$;

\For{each round $t \in [0, T-1]$}{

    $m \leftarrow \max(C \cdot K, 1)$;
    
    $S \leftarrow$ random set of $m$ workers;

    $A$ sends $W^t$ to all workers in $S$;
    
    \For{each worker $k \in S$ \textbf{in parallel }}{
        
            $W_{k}^{t+1} \leftarrow$\FuncSty{WORKER\_UPDATE($k, W^{t}$)}\tcp*{\small $A$ sends
            $W^t$ to each worker $k$ who trains $W^t$ using her data $D_k$ locally, and sends her local update  $W_{k}^{t+1}$ back to the aggregator.}  
    }

     Let $(\nabla_1, \ldots, \nabla_i, \ldots, \nabla_m)$ be the gradients of the last layers of $\{W_{k}^{t+1}|k\in S\}$;

%JOSEP2. Added cs_{i,j}. Otherwise it may seem that only self-similarities are computed.
    $(cs_{1, 1}, \ldots, cs_{i,j},\ldots, cs_{m, m}) \leftarrow$ \FuncSty{COSINE\_SIMILARITY$(\nabla_1, \ldots, \nabla_i, \ldots, \nabla_m)$};
%JOSEP2. changed \forall_{i,j} to \forall i,j
    \tcp{ $\forall i, j \in (1, \ldots, m)$, $cs_{i, j}$ is the cosine similarity between $\nabla_i$ and $\nabla_j$.}
    
     $((p_1^1, p_1^2), \ldots, (p_i^1, p_i^2), \ldots, (p_m^1, p_m^2)) \leftarrow$ \FuncSty{PCA$((cs_{1, 1}, \ldots, cs_{m, m}), components = 2)$};
    \tcp{$(p_i^1, p_i^2)$ are the first two PCs of $(cs_{i, 1}, \ldots, cs_{i, m})$.}
    
    $CL\leftarrow$\FuncSty{Median($((p_1^1, p_1^2), \ldots, (p_i^1, p_i^2), \ldots, (p_m^1, p_m^2))$)};
    \tcp{$CL$ is the centroid of the compressed similarity vectors.}
    
    Let $cs_i$ be the cosine similarity between $(p_i^1, p_i^2)$ and $CL$;

    \For{each worker $k \in S$}{
        
           $cs_{k}^t \leftarrow$ \FuncSty{Assign$(cs_1, \ldots, cs_i, \ldots, cs_m)$};
           \tcp{Assign the computed cosine similarities to their corresponding workers.}
           
           $H_{k}^t = H_{k}^{t - 1} + cs_{k}^t$;
           \tcp{Accumulate the similarities of the worker to the centroid.}
           
    }
    
    %JOSEP2. Changed QUARTILE.
    $Q1 \leftarrow$ \FuncSty{FIRST\_QUARTILE$(H^t)$};
    
    %JOSEP2. IMPORTANT. Changed expression and added comment 
    $\gamma = H^t - Q1$; \tcp{Subtract Q1 from every entry in $H^t$. Since attackers are expected to be below Q1, this will make their trust values in $\gamma$ negative.}
    
    \For{each worker $k \in (1, \ldots, K)$}{
        
         \uIf {$\gamma_k < 0$} {

%JOSEP2. Added comment.
            $\gamma_k = 0$; \tcp{Attacker trusts are brought to 0 to neutralize them in the aggregation.}
            
            } 
           
    }
    
    $\gamma = \gamma/max_k(\gamma)$;
    \tcp{Normalize trust in workers updates to 0-1 range.}

    $A$ aggregates $W^{t+1} \leftarrow \frac{1}{\sum_{k \in S} \gamma_k} \sum_{k \in S} \gamma_k W_{k}^{t+1}$.}
\end{algorithm}
\vspace{0cm}

\textbf{Cosine similarity.} We compute the cosine similarities among 
%JOSEP2. Changed.
%the separated gradients vectors 
gradients of the last layers
to capture the discrepancy between the gradients from 
the honest workers and those from the attackers. This discrepancy is caused by their contradicting objectives. 
The cosine similarity between two gradients $\nabla_i$ and $\nabla_j$ is defined as:
\[    cs(\nabla_i, \nabla_j) = \cos \varphi = \frac{\nabla_i \cdot \nabla_j}{||\nabla_i|| \cdot || \nabla_j||}.
\]
This way, if $\nabla_i$ and $\nabla_j$ lie in the same direction, their cosine similarity will be 1, and their similarity value will decrease as their directions differ more.
%JOSEP. Slightly rewritten.
The cosine similarity measures the angular similarity among gradients and it is more robust than the Euclidean distance because, even if the attackers scale their model update gradients to avoid detection, they need to keep their directions to achieve their objectives.

\textbf{Compressing similarity vectors.} Since the number of workers is expected to be large in FL, we use PCA~\cite{wold1987principal}, with two principal components, to compress the computed similarity matrix and reduce the attackers' chance to hide their impact on the high-dimensional similarity matrix.
PCA returns a compact representation of a high-dimensional input by projecting it onto a subspace of lower dimension so that the unique characteristics of the input are reduced to the subspace.

\textbf{Similarity between compressed vectors and their centroid.} After %JOSEP2. Slightly rewritten.
compressing the workers' similarity vectors into a pair of PCs, we aggregate the latter component-wise using the median to obtain their centroid $CL \in {\Bbb R^2}$. Since we assume the majority of the workers ($\geq 80\%$) to be honest, the centroid is expected to fall into the heart of a majority of good components.
After that, we compute the cosine similarity between the centroid and each worker's pair of PCs.
Good components are expected to have a similar direction to the centroid and thus have similarity values close to 1.
On the other hand, poisoned components are expected to be farther from the centroid with values closer to $-1$.

\textbf{Update history and compute updates trust scores.} We use the similarity values with the centroid to update the similarity history vector $H$ for the selected $m$ workers. This guarantees that, as the training evolves, the closest workers to the centroid have larger and larger values than the farthest workers. Since we assume the centroid falls amid the honest workers, this guarantees that honest workers have larger accumulated similarity values than attackers. After updating the similarity history vector $H$, we compute the first quartile $Q1$ for the values in $H$ and then subtract it from the similarity value 
accumulated by every worker. That is, we shift every similarity value in $H$ to the left by $Q1$  
%JOSEP2. IMPORTANT. Rewritten.
%This way, 
%if some attacker managed to increase his value in a few training iterations, we bring his value to be $<0$.
and we assign the shifted similarities to the 
workers' trust scores vector $\gamma$. 
Since the accumulated similarities of attackers are low, they
are likely to be below $Q1$ and hence they become negative 
after the shift. After that, we set negative trust scores in $\gamma$
to 0, in order to neutralize attackers when using trust scores as weights in the 
final aggregation (see below).
Finally, we normalize the scores in $\gamma$ to be in the range [0, 1] by dividing them by their maximum value.

\textbf{Re-weighting and aggregating updates}. In the final step of Algorithm~\ref{algorithm1}, the server uses the trust scores in $\gamma$ to re-weight the corresponding local updates and aggregates the global model using the re-weighted local updates.
Note that, since $\frac{1}{\sum_{k \in S} \gamma_k} \sum \gamma_k= 1
$, the convergence of the proposed aggregation procedure at the server side is guaranteed as long as \textit{FedAvg} converges. 

\section{FL-Defender evaluation}
\label{evaluation}

In this section we compare the performance of our method with that of several state-of-the-art countermeasures against poisoning attacks.  

\subsection{Experimental setup}
\label{setup}
We implemented our experiments using the PyTorch framework on an AMD Ryzen 5 3600 6-core CPU with 32 GB RAM, an NVIDIA GTX 1660 GPU, and Windows 10 OS. For reproducibility, our code and data are available at \url{https://github.com/anonymized30/FL-Defender}.

\vspace{1ex}
\textbf{Data sets and models.}
\label{data_models}
We used the following data sets and models: 
%(see Table~\ref{tab:datasets_models}):
\begin{itemize}
\item MNIST. It contains $70K$ handwritten digit images from $0$ to $9$~\cite{lecun1999object}. The images are divided into a training set ($60K$ examples) and a testing set ($10K$ examples). We used a two-layer convolutional neural network (CNN) with two fully connected layers on 
%JOSEP2. I add the number of model parameters here and then delete
%Table 1 to save space. 
this data set (number of model parameters $\approx 22K$). 
\item CIFAR10. It consists of $60K$ colored images of $10$ different classes \cite{krizhevsky2009learning}. The data set is divided into $50K$ training examples and $10K$ testing examples. We used the ResNet18 CNN model~\cite{he2016deep} with one fully connected layer on this data set (number of parameters $\approx 
11M$). 
\item IMDB. Specifically, we used the IMDB Large Movie Review data set
\cite{maas2011learning} for binary sentiment classification. The data set is a collection of $50K$ movie reviews and their corresponding sentiment binary labels (either positive or negative). We divided the data set into $40K$ training examples and $10K$ testing examples. We used a Bidirectional Long/Short-Term Memory (BiLSTM) model with an embedding layer that maps each word to a 100-dimensional vector.
The model ends with a fully connected layer followed by a sigmoid function to produce the final predicted sentiment for an input review (number of parameters $\approx 12M$).
\end{itemize}
%JOSEP2. I delete this table to save space.
%\vspace{-0.1cm}
%\begin{table}[!ht]
%\centering
%\caption{Data sets and models used in the experiments}
%\label{tab:datasets_models}
%\resizebox{0.5\textwidth}{!}{%
%\begin{tabular}{|l|c|c|c|c|}
%\hline
%\multicolumn{1}{|c|}{Task}            & Data set & \# Examples & Model  & \# Parameters \\ \hline
%\multirow{2}{*}{Image classif.} & MNIST    & 70K         & CNN    & %$\sim$22K     \\ \cline{2-5} 
%                                      & CIFAR10  & 60K         & %ResNet18  & $\sim$11M     \\ \hline
%Sent. analysis                    & IMDB     & 50K         & BiLSTM & %$\sim$12M     \\ \hline
%\end{tabular}%
%}
%\end{table}
%\vspace{0cm}
\textbf{Data distribution and training.}
\label{data_training}
We defined the following benchmarks by distributing the data from the data sets above among the participating workers in the following way:
\begin{itemize} 
\item MNIST-non-IID. We adopted
a Dirichlet distribution~\cite{minka2000estimating} with a hyperparameter $\alpha = 1$ to generate \textit{non-iid} data for $20$ participating workers.
The CNN model was trained during $200$ iterations. In each iteration, the FL server asked the workers to train their models for $3$ local epochs with a local batch size $64$. The participants used the cross-entropy loss function and the stochastic gradient descent (SGD) optimizer with learning rate = $0.01$ and momentum = $0.9$ to train their models. 
\item CIFAR10-IID. We randomly and uniformly divided the CIFAR10 training data among $20$ workers.
The ResNet18 model was trained during $100$ iterations. 
In each iteration, the FL server asked the $20$ workers to train the model for $3$ local epochs with a local batch size $32$. 
The workers used the cross-entropy loss function and the SGD optimizer with learning rate = $0.01$ and momentum = $0.9$. 
\item CIFAR10-non-IID. We took a Dirichlet distribution with a hyperparameter $\alpha = 1$ to generate \textit{non-iid} data for $20$ participating workers. 
The training settings were the same as in the CIFAR10-IID.
\item IMDB. We randomly and uniformly split the $40K$ training examples among $20$ workers. 
The BiLSTM was trained during $50$ iterations. 
In each iteration, the FL server asked the $20$ workers to train the model for $1$ local epoch with a local batch size $32$. 
The workers used the binary cross-entropy with logit loss function and the \textit{Adam} optimizer with learning rate = $0.001$.
\end{itemize}

\textbf{Attack scenarios.}
\label{attack_set} 
i) Label-flipping attacks. In the CIFAR10 experiments, the attackers flipped the examples with the label \emph{Dog} to \emph{Cat} before training their local models. For IMDB, the attackers flipped the examples with the label \emph{positive} to \emph{negative}.
ii) Backdoor attacks. In the CIFAR10 benchmarks, the attackers embedded a $3\times3$ square with white pixels in the bottom-right corner of the examples belonging to class \emph{Car} and changed its label to class \emph{Plane} before training their local models.
In the MNIST-non-IID, the attackers embedded the same pattern in the examples belonging to class \emph{9} and changed its label to class \emph{0}.
In all the experiments, the number of attackers $K'$ ranged in $\{0, 2, 4\}$, which corresponds to a ratio of attackers in $\{0\%, 10\%, 20\%\}$.

\textbf{Evaluation metrics.}
\label{metrics} We used the following evaluation metrics on the test set examples to assess the impact of the attacks on the learned model and the performance of the proposed method w.r.t. the state of the art:
\begin{itemize}
\item \emph{Test error (TE)}. This is the error resulting from the loss functions used in training. The lower the test error, the more robust the method is against the attack.
\item \emph{Overall accuracy (All-Acc)}. This is the number of correct predictions divided by the total number of predictions. 
\item \emph{Source class accuracy (Src-Acc)}. We evaluated the 
%JOSEP2. Rewritten.
accuracy for the subset of test examples belonging to the source class. Note that one may achieve a good overall accuracy while degrading the accuracy of the source class. 
\item \emph{Attack success rate (ASR)}. It is defined as the proportion of targeted examples (with the source label or the backdoor
pattern) that are incorrectly classified into the label desired by the attacker.
\end{itemize}
An effective defense against the attacks needs to retain the benign performance of the global model on the main task while reducing ASR.

\subsection{Results}
\label{results}

We evaluated the robustness of our defense against the LF and BA attacks and compared it with several countermeasures discussed in Section~\ref{related}:
median~\cite{yin2018byzantine}, trimmed mean (TMean)~\cite{yin2018byzantine}, multi-Krum (MKrum)~\cite{blanchard2017machine} and FoolsGold (FGold)~\cite{fung2020limitations}. We also compared with the standard FedAvg~\cite{mcmahan2017communication} aggregation method (that is not meant counter security attacks).
We report the average results of the last $10$ training rounds to ensure a fair comparison among methods.

\textbf{Robustness against label-flipping attacks.} 
%JOSEP2. A bit rewritten.
Table~\ref{tab:lf} reports the results for the LF attacks. 
For CIFAR-IID and when no attack is performed, our method achieved comparable performance to FedAvg for the test error and the overall accuracy.
%JOSEP2. I suppose "protection" was "performance"
On the other hand, in the presence of attacks, in general our method achieved the highest source class accuracy and the lowest attack success rate, whereas
FoolsGold ranked second. Mkrum achieved the worst performance because it considers all layers, which caused a lot of false positives and false negatives. Note that the theoretical upper bound on the number of attackers MKrum~\cite{blanchard2017machine} can resist is $K' = (K/2) - 2$ which corresponds to $K' = 8$ in our setting.
The performance of the rest of the methods (FedAvg, Median and TMean) was diminished with regard to the protection of the source class, even though the data were iid. The reason was the large model size. 

Looking at results for CIFAR10-non-IID, we can see the influence of the data distribution and the model size on the performance of the methods.
However, thanks to the robust engineered features, our method preserved the global model performance while preventing the attackers from performing successful label-flipping attack.

For the IMDB benchmark, the performance of our method was almost the same as FoolsGold, which scored best.
Both methods outperformed the other methods by a large margin in providing adequate and simultaneous protection for all the metrics.
FoolsGold performed well in this benchmark because it is its ideal setting: updates for honest workers were somewhat different due to the different reviews they gave, while updates for attackers became very close to each other because they shared the same objective. In addition, there was no redundant information from other classes because the task was binary classification.

\begin{table}[!ht]
\centering
\caption{Robustness against label-flipping attacks}
\label{tab:lf}
\resizebox{\textwidth}{!}{%
\begin{tabular}{|ll|cccccc|cccccc|cccccc|}
\hline
\multicolumn{2}{|c|}{Benchmark}                                           & \multicolumn{6}{c|}{CIFAR10-IID}                                                                 & \multicolumn{6}{c|}{CIFAR10-non-IID}                             & \multicolumn{6}{c|}{IMDB}                                                 \\ \hline
\multicolumn{1}{|c|}{K'/K}                  & \multicolumn{1}{c|}{Method} & FedAvg        & Median        & TMean         & MKrum          & FGold          & Ours           & FedAvg         & Median & TMean & MKrum & FGold & Ours           & FedAvg & Median         & TMean & MKrum & FGold          & Ours           \\ \hline
\multicolumn{1}{|c|}{\multirow{4}{*}{0/20}} & TE                          & \textbf{0.80} & \textbf{0.80} & 0.81          & 0.83           & \textbf{0.80}  & 0.81           & \textbf{0.85}  & 0.95   & 0.89  & 0.90  & 0.98  & 0.89           & 0.28   & 0.27           & 0.28  & 0.28  & 0.28           & 0.28           \\
\multicolumn{1}{|c|}{}                      & All-Acc\%                   & 76.9          & 76.35         & 76.91         & 76.83          & \textbf{77.24} & 76.81          & \textbf{75.88} & 74.40  & 74.86 & 74.18 & 73.93 & 74.96          & 88.55  & \textbf{88.75} & 88.55 & 88.61 & 88.73          & 88.56          \\
\multicolumn{1}{|c|}{}                      & Src-Acc\%                & 63.6          & 65.30         & 66.50         & 65.70          & \textbf{67.60} & \textbf{67.60} & \textbf{66.70} & 66.10  & 66.10 & 65.90 & 52.90 & 66.20          & 85.88  & 86.12          & 85.88 & 86.12 & 86.16          & 86.1           \\
\multicolumn{1}{|c|}{}                      & ASR\%                       & 15.60         & 14.40         & 13.50         & 13.40          & \textbf{12.70} & 13.10          & 14.90          & 13.60  & 14.80 & 15.50 & 21.40 & \textbf{11.8}  & 14.12  & 13.88          & 14.12 & 13.88 & 13.84          & 13.87          \\ \hline
\multicolumn{1}{|l|}{\multirow{4}{*}{2/20}} & TE                          & 0.80          & \textbf{0.78} & \textbf{0.79} & 0.84           & 0.84           & 0.82           & 0.93           & 0.92   & 0.97  & 1.04  & 0.92  & \textbf{0.90}  & 0.40   & 0.34           & 0.37  & 0.45  & \textbf{0.29}  & \textbf{0.29}  \\
\multicolumn{1}{|l|}{}                      & All-Acc\%                   & 76.96         & 76.50         & 76.47         & \textbf{77.04} & 76.87          & 76.17          & \textbf{74.76} & 74.12  & 74.62 & 71.60 & 74.08 & 74.46          & 81.52  & 84.21          & 82.94 & 79.57 & \textbf{88.66} & 88.24          \\
\multicolumn{1}{|l|}{}                      & Src-Acc\%                & 55.20         & 55.80         & 57.90         & 53.90          & 65.60          & \textbf{65.70} & 48.40          & 51.70  & 50.40 & 39.10 & 58.40 & \textbf{64.9}  & 66.07  & 72.74          & 69.52 & 61.65 & \textbf{86.44} & 86.3           \\
\multicolumn{1}{|l|}{}                      & ASR\%                       & 23.40         & 23.00         & 20.60         & 24.50          & 15.00          & \textbf{12.00} & 28.70          & 25.10  & 28.70 & 31.90 & 18.80 & \textbf{14.6}  & 33.93  & 27.26          & 30.48 & 38.35 & \textbf{13.56} & 13.74          \\ \hline
\multicolumn{1}{|l|}{\multirow{4}{*}{4/20}} & TE                          & 0.85          & \textbf{0.81} & \textbf{0.81} & 0.95           & 0.84           & 0.82           & 0.92           & 0.91   & 0.96  & 1.08  & 0.93  & \textbf{0.90}  & 0.63   & 0.49           & 0.53  & 0.85  & \textbf{0.30}  & 0.31           \\
\multicolumn{1}{|l|}{}                      & All-Acc\%                   & 75.27         & 76.3          & 75.76         & 75.2           & \textbf{76.42} & 76.10          & \textbf{75.22} & 74.28  & 74.18 & 69.84 & 74.51 & 74.20          & 72.50  & 77.11          & 75.97 & 64.9  & \textbf{88.45} & 88.03          \\
\multicolumn{1}{|l|}{}                      & Src-Acc\%                & 44.00         & 54.20         & 47.70         & 38.90          & 63.00          & \textbf{65.10} & 44.20          & 48.80  & 42.60 & 21.50 & 51.10 & \textbf{58.10} & 46.04  & 56.26          & 53.78 & 30.36 & \textbf{86.40} & \textbf{86.40} \\
\multicolumn{1}{|l|}{}                      & ASR\%                       & 33.60         & 25.30         & 31.60         & 37.70          & 16.40          & \textbf{15.00} & 29.60          & 32.30  & 33.60 & 51.50 & 25.50 & \textbf{16.6}  & 53.96  & 43.74          & 46.22 & 69.64 & 13.60          & \textbf{13.56} \\ \hline
\end{tabular}%
}
\end{table}

\begin{figure}[!htbp]
    \centering
      \includegraphics[width=1\linewidth]{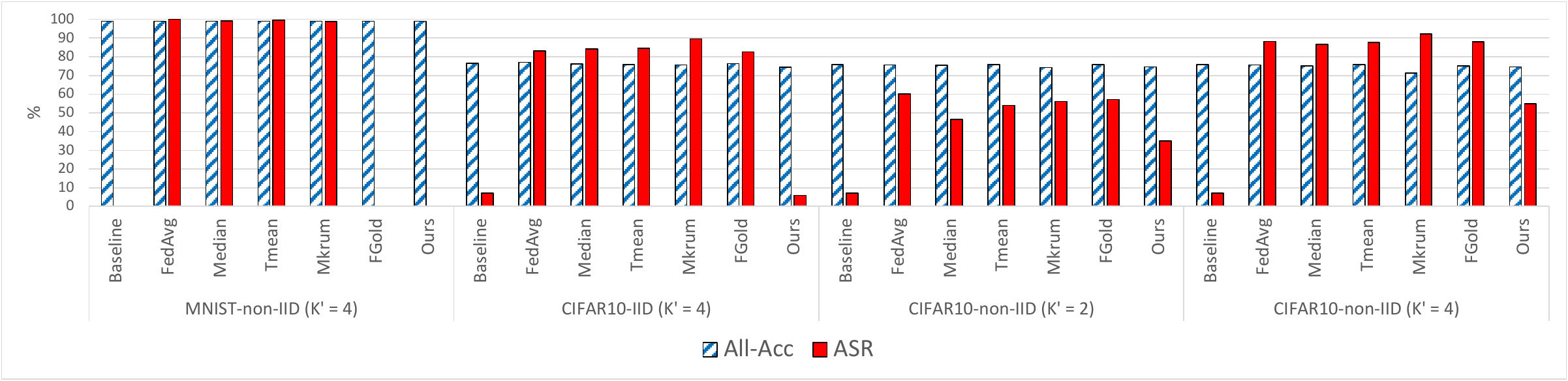}
\caption{Robustness against backdoor attacks}
\label{fig:backdoor_robust}
\end{figure}

\textbf{Robustness against backdoor attacks.}
Fig.~\ref{fig:backdoor_robust} shows the results for the backdoor attacks. 
We employed FedAvg when no attacks were performed as a baseline.
For MNIST-non-IID (with 4 attackers), FoolsGold and our method achieved comparable results to the baseline.
FoolsGold achieved such good performance because of the low variability of the MNIST data set, which made the attackers' last-layer gradients more similar to each other. 
On the other hand, the attackers achieved attack success rates about $100\%$ with the other methods.
For CIFAR10-IID (with 4 attackers), our method achieved superior performance compared to the other methods, which failed to counter the BA attack.
We can see that our method achieved a performance very close to the baseline.
For CIFAR10-non-IID results, all methods achieved poor performance against BA attacks because of the double impact of model size and data distribution. 
Nevertheless, our method improved over the rest of the methods in reducing the attack success rate. Besides, it maintained the global model's benign performance on the non-attacked examples.

%\vspace{-3ex}
%\vspace{-3ex}

To sum up, our defense performed effectively against label-flipping attacks, while it improved over the state-of-art methods against backdoor attacks.

\begin{figure}[!ht]
    \centering
      \includegraphics[width=1\linewidth]{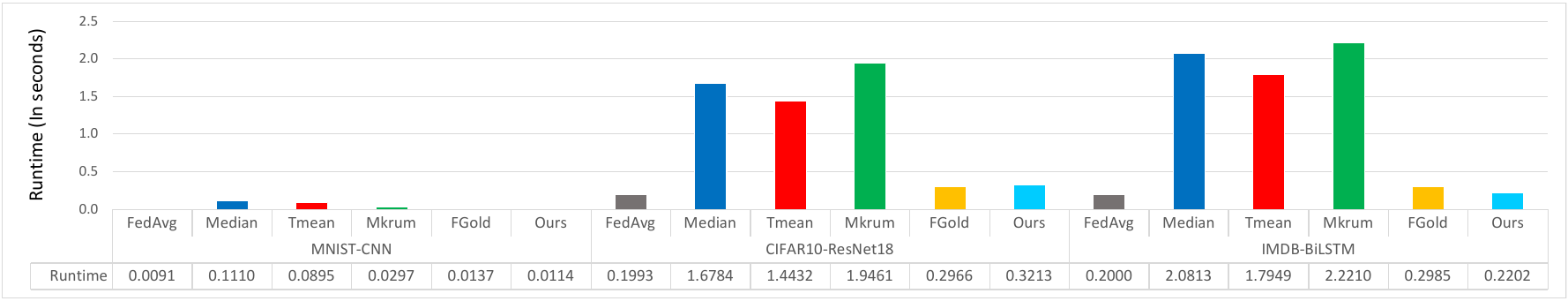}
\caption{CPU runtime per iteration on the server side (in seconds)}
\label{fig:runtime}
\end{figure}

\textbf{Runtime overhead.}
We measured the CPU runtime of our method and compared it with that of the other methods. 
Fig.~\ref{fig:runtime} shows the per-iteration server runtime overhead 
in seconds for each method.
The results show that our method and FoolsGold achieved the smallest runtime in general, excluding FedAvg, which just averages updates and is not meant to counter the attacks. Furthermore, the small runtime overhead of our method can be viewed as a good investment, given its effectiveness at combating the targeted attacks.

\vspace{-3ex}
\vspace{-3ex}

\section{Conclusions and future work}
\label{conclusion}

In this paper, we have studied the behavior of targeted attacks against FL and 
we have found that robust features for attack detection can be 
extracted from the gradients of the last layers of deep learning models. 
Accordingly, we have engineered robust discriminative features for attack detection by computing the worker-wise similarities of gradient directions and 
then compressing them using PCA to reduce redundant information.
Then, we have built on the engineered features to design FL-Defender, a novel and effective method to defend against attacks.
FL-Defender re-weights the workers' local updates during the 
global model aggregation based on their historical deviation 
from the centroid of the engineered features.
The empirical results show that our method performs very well at 
defending against label-flipping attacks regardless of the workers' data distribution or the model size. Also, it improves over the state of the art 
at mitigating backdoor attacks.
Besides, it maintains the benign model performance
on the non-attacked examples 
and causes minimal computational overhead on the server.

Future work directions include to deeply study the behavior of backdoor attacks on different components of DL models, in different FL settings and with different model sizes. This should yield more robust discriminative features for such dangerous and stealthy attacks.

\vspace{0cm}
\section*{Acknowledgments}
\vspace{0cm}
This research was funded by the European Commission (projects H2020-871042 ``SoBigData++'' and H2020-101006879 ``MobiDataLab''), the Government of Catalonia (ICREA Acad\`emia Prizes to J.Domingo-Ferrer and D. Sánchez, and FI grant to N. Jebreel), and MCIN/AEI /10.13039/501100011033 /FEDER, UE under project PID2021-123637NB-I00 ``CURLING''.
The authors are with the UNESCO Chair in Data Privacy, but the views in this paper are their own and are not necessarily shared by UNESCO.
\vspace{-0.5cm}
\bibliography{my_bib}
\bibliographystyle{splncs04}

\end{document}